\documentclass[acmsmall, noacm]{acmart}
\AtBeginDocument{%
  }

\setcopyright{none}

\usepackage{tabularx}
\usepackage{array}
\usepackage{xcolor}
\usepackage{booktabs}
\usepackage{threeparttable}
\usepackage{longtable,ragged2e}

\renewcommand\footnotetextcopyrightpermission[1]{} 
\setcopyright{none} 
\acmVolume{}
\acmNumber{}
\acmArticle{}

\begin{document}

\title{Beyond the Uncanny Valley: A Mixed-Method Investigation of Anthropomorphism in Protective Responses to Robot Abuse}

\author{Fan Yang}
\authornote{Corresponding Author}
\email{trovato@corporation.com}
\affiliation{%
  \institution{University of South Carolina}
  \city{Dublin}
  \state{Ohio}
  \country{USA}
}

\author{Lingyao Li}
\affiliation{%
  \institution{University of South Florida}
  \city{Tampa}
  \country{USA}}
\email{lingyaol@usf.edu}

\author{Yaxin Hu}
\affiliation{%
  \institution{University of Wisconsin--Madison}
  \city{Madison}
  \country{USA}}
\email{yaxin.hu@wisc.edu}

\author{Michael D. Rodgers}
\affiliation{%
  \institution{University of South Carolina}
  \city{Columbia}
  \country{USA}}
\email{mr182@email.sc.edu}

\author{Renkai Ma}
\affiliation{%
 \institution{University of Cincinnati}
 \city{Cincinnati}
 \state{Ohio}
 \country{USA}}
 \email{renkai.ma@uc.edu}

\newcommand{\YF}[1]{\textcolor{magenta}{Fan: #1}}
\newcommand{\RK}[1]{\textcolor{blue}{[Renkai: #1]}}
\newcommand{\LY}[1]{\textcolor{orange}{[Lingyao: #1]}}

\renewcommand{\shortauthors}{Yang et al.}

\begin{abstract}
    Robots with anthropomorphic features are increasingly shaping how humans perceive and morally engage with them. Our research investigates how different levels of anthropomorphism influence protective responses to robot abuse, extending the Computers as Social Actors (CASA) and uncanny valley theories into a moral domain. In an experiment, we invite 201 participants to view videos depicting abuse toward a robot with low (Spider), moderate (Two-Foot), or high (Humanoid) anthropomorphism. To provide a comprehensive analysis, we triangulate three modalities: self-report surveys measuring emotions and uncanniness, physiological data from automated facial expression analysis, and qualitative reflections. Findings indicate that protective responses are not linear. The moderately anthropomorphic Two-Foot robot, rated highest in eeriness and ``spine-tingling'' sensations consistent with the uncanny valley, elicited the strongest physiological anger expressions. Self-reported anger and guilt are significantly higher for both the Two-Foot and Humanoid robots compared to the Spider. Qualitative findings further reveal that as anthropomorphism increases, moral reasoning shifts from technical assessments of property damage to condemnation of the abuser's character, while governance proposals expand from property law to calls for quasi-animal rights and broader societal responsibility. These results suggest that the uncanny valley does not dampen moral concern but paradoxically heightens protective impulses, offering critical implications for robot design, policy, and future legal frameworks.
\end{abstract}




\maketitle

\section{INTRODUCTION}
The Computers as Social Actors (CASA) paradigm ~\cite{10.1145/191666.191703, nass2000machines} has been extensively examined in numerous empirical studies across different contexts ~\cite{gambino2020building,katagiri2001cross}. From computers ~\cite{lee2010trust, nass1993voices} to AI ~\cite{park2021computers, xu2022deep,seok2025emotions, gu2024exploring, zhao2025tailoring}, and to robots ~\cite{lee2005can, xu2023media, xu2019first}, scholars have found that humans are inclined to apply social rules (e.g., etiquette, reciprocity, stereotypes) to social agents, especially when they resemble human characteristics or display social cues ~\cite{lombard2021social}, a phenomenon known as anthropomorphism ~\cite{gong2008social, nass2000machines, lee2010triggers, kim2012anthropomorphism}. However, extensive human-robot interaction (HRI) research also suggests a potential caveat: social responses to robots do not always increase linearly with anthropomorphism. Instead, near-human appearance can paradoxically trigger aversion and discomfort and subsequently inhibit engagement with a robot, a hypothesis known as the uncanny valley effect ~\cite{mori2012uncanny, macdorman2006subjective, zhang2020literature}. 

While existing literature has shown mixed evidence, with both support for and challenges to the existence of CASA ~\cite{heyselaar2023casa,lee2010triggers, antos2011influence, gambino2020building} and the uncanny valley ~\cite{poliakoff2013can, diel2021meta, weisman2024updating, urgen2018uncanny, zlotowski2015persistence}, the majority of existing HRI research has examined them primarily in benign social interaction contexts (e.g., engaging conversations, performing collaborative tasks)~\cite{chiou2020we, seok2025emotions, de2024siot}. This raises a critical question: Can the uncanny valley effect also impact how we apply moral rules to robots when they are mistreated?

By examining protective responses to robots in abusive rather than neutral or cooperative interactions, this study investigates whether the psychological mechanisms predicted by the uncanny valley hypothesis can also govern moral concern in a similar fashion. While people may avoid interacting with uncanny robots due to discomfort, they might still feel morally obligated to protect them from harm. Alternatively, the uncanny valley effect might be so powerful that it suppresses protective impulses even when robots are clear victims of mistreatment.

Utilizing a mixed-methods experimental research design that triangulates three distinct data modalities—objective physiological measures, self-report survey responses, and qualitative open-ended reflections, this study provides a comprehensive empirical examination of how anthropomorphism, a core concept in both CASA and the uncanny valley effect, influences protective responses to robot abuse. Findings of this research study directly extend CASA and uncanny valley into a novel context of robot abuse and advance theoretical understanding of how anthropomorphism impacts moral considerations towards robots. In addition, these findings also have critical practical implications for robot design, deployment policies, and legal frameworks, informing how robots should be designed to elicit appropriate moral consideration, particularly in vulnerable contexts such as public service ~\cite{shiomi2025robot, shum2024kicking}, education ~\cite{tan2018inducing}, and hospitality ~\cite{sun2026frontline} where robots may face mistreatment. 

Building on the existing CASA \cite{nass2000machines, nass1993voices} and uncanny valley \cite{mori2012uncanny, macdorman2006subjective} literature, our study employs a mixed-method experimental design combining quantitative, physiological, and qualitative data to investigate the role of anthropomorphism - a central concept in both CASA and uncanny valley - in shaping up protective responses to robot abuse. Understanding how anthropomorphism influences moral consideration critically extends both CASA paradigm and uncanny valley hypothesis beyond their currently tested boundaries and addresses practical concerns about robot abuse and its potential to shape societal norms. As such, we ask:

\begin{itemize}
    \item \textbf{RQ1}: How do individual self-report reactions to robot abuse vary by a robot's anthropomorphic level?
    \item \textbf{RQ2}: How do individual physiological reactions to robot abuse vary by a robot's anthropomorphic level?
    \item \textbf{RQ3}: How do individual views on robot abuse and protection vary by a robot's anthropomorphic level, if any?
\end{itemize}

\section{RELATED WORK}
\subsection{Robots as Social Actors}

A central debate about CASA resides in whether humans apply social rules to intelligent agents mindlessly—as automatic responses—or mindfully through conscious anthropomorphism ~\cite{kim2012anthropomorphism, xu2023media}. While the original CASA assumed mindless application of social scripts regardless of anthropomorphic cues ~\cite{nass2000machines}, significant research has shown that anthropomorphism plays a crucial, albeit complex,  role in driving social responses to technologies ~\cite{lombard2021social}. For instance, while Lee (2010) found that anthropomorphic cartoon characters enhanced social attractiveness and trustworthiness of computers ~\cite{lee2010triggers}, the other study demonstrated significant influences of anthropomorphic design cues on agent perceptions ~\cite{araujo2018living}.

Further scholarly inquiries discover that anthropomorphism can occur both mindlessly and mindfully, suggesting these are not mutually exclusive mechanisms ~\cite{kim2012anthropomorphism}. In the attempt to unify the two explanatory mechanisms of CASA - mindless and mindful anthropomorphism, Lombard and Xu (2021) propose that mindless anthropomorphism better explains social responses when technologies display high-quality and high-quantity social cues, such as highly human-like robots ~\cite{phillips2018human, christou2020tourists}, where users experience intuitive, spontaneous responses to natural and powerful cues. Conversely, mindful anthropomorphism better explains strong social responses to technologies with limited cues, where individuals consciously project human qualities onto non-humanlike objects. Despite the two distinct mechanisms, an overarching theme that emerges from the existing CASA literature is that anthropomorphism fundamentally influences how users transfer social rules from human-human to human-technology interactions, including how they interact with robots as social actors ~\cite{fink2012anthropomorphism, zlotowski2015anthropomorphism}.

\subsection{Uncanny Effect in Human-Robot Interaction}
The uncanny valley hypothesis \cite{mori2012uncanny,brenton2005uncanny} gives anthropomorphism an interesting twist: rather than assuming a linear relationship where more human-like features consistently promote favorable social responses, it predicts a nonlinear pattern with a dip where near-humanness triggers aversion and turns off acceptance \cite{macdorman2016reducing, katsyri2015review}. The primary psychological explanation for the existence of the uncanny valley centers on expectation violation and perceptual mismatch theories, which propose that human-like appearance generates expectations for biological behavior that mechanical movement violates, triggering cognitive dissonance and discomfort \cite{saygin2012thing, urgen2018uncanny}. 

However, growing HRI literature has pointed out a series of moderating factors that could amplify, dilute, or even nullify the uncanny valley effect. For example, several individual differences, such as neuroticism and religious beliefs, will significantly predict a person's susceptibility to the uncanny valley \cite{macdorman2015individual, feng2018uncanny}. Context also matters: repeated exposure reduces initial eeriness, behavioral cues can override appearance-based judgments for machine-like robots but not androids already in the valley, and cultural background shapes baseline acceptance of anthropomorphic agents \cite{zlotowski2015anthropomorphism}. Recent meta-analyses \cite{diel2021meta, katsyri2015review} suggest the phenomenon may not necessarily be a ``valley'' \cite{bartneck2007uncanny} but rather negative responses conditioned upon specific features mismatch, individual characteristics and contextual factors.

\subsection{Perceiving and Responding to Robot Abuse}

Robot abuse, defined as actions by users against robots that resemble human-on-human aggression ~\cite{rezzani2025robot, gallego2019parent, bartneck2008exploring, shiomi2025robot, yamada2020escalating, keijsers2018mindless, rezzani2023space}, has emerged as a concerning phenomenon in human-robot interaction. As robots become increasingly integrated into public spaces and daily life, documented instances of unprovoked aggression toward these machines raise important questions about human moral reasoning, social behavior, and the boundaries of ethical consideration ~\cite{bartneck2008exploring, bartneck2020morality, yamada2025development}. Such abuse ranges from verbal harassment ~\cite{sanchez2018preliminary, chin2020empathy} and physical strikes ~\cite{sparrow2016kicking, shum2024kicking} to deliberate destruction ~\cite{rossmy2020punishable, bartneck2007kill}, often occurring despite the absence of any threat or provocation from the robot itself.

Scholarly works have increasingly focused on examining attitudes towards and (un)acceptance of robot abuse. Building on the CASA paradigm, scholars have argued that people, theoretically, can engage in aggressive behaviors toward robots that mirror patterns of human-to-human aggression~\cite{bartneck2008exploring}. However, willingness to mistreat robots is significantly influenced by their design~\cite{rossmy2020punishable}, with previous literature demonstrating a strong linear relationship between anthropomorphic features and empathy toward robots ~\cite{riek2009anthropomorphism, riek2009empathizing}. Studies have shown that anthropomorphic behavioral traits—agreeableness and intelligence—significantly reduced participants' willingness to turn off a robot ~\cite{bartneck2007daisy, horstmann2018robot}. 

In the specific context of robot abuse, Bartneck and Hu (2008) found that intelligence significantly moderates people's destructive behaviors (e.g., killing) towards robots. Functional magnetic resonance imaging (fMRI) studies offer neuroscientists evidence for this, with research suggesting that our neural activation when witnessing robots being abused mirrors the patterns activated when seeing a human being treated violently ~\cite{keijsers2018mindless}, though the degree of activation is significantly higher for humans than for robots ~\cite{rosenthal2013neural, bartneck2008exploring}. It is, therefore, reasonable to predict that robots that can better mimic human traits, whether in appearance, personality, or behavior, can elicit stronger emotional and protective responses when subjected to abuse. However, it remains unknown whether the uncanny valley effect, typically associated with increasing anthropomorphism, might disrupt this linear relationship between perceived anthropomorphism and protective responses to robot abuse.

\section{METHODS}

\begin{figure*}[t]
    \centering
    \includegraphics[width=1\textwidth]{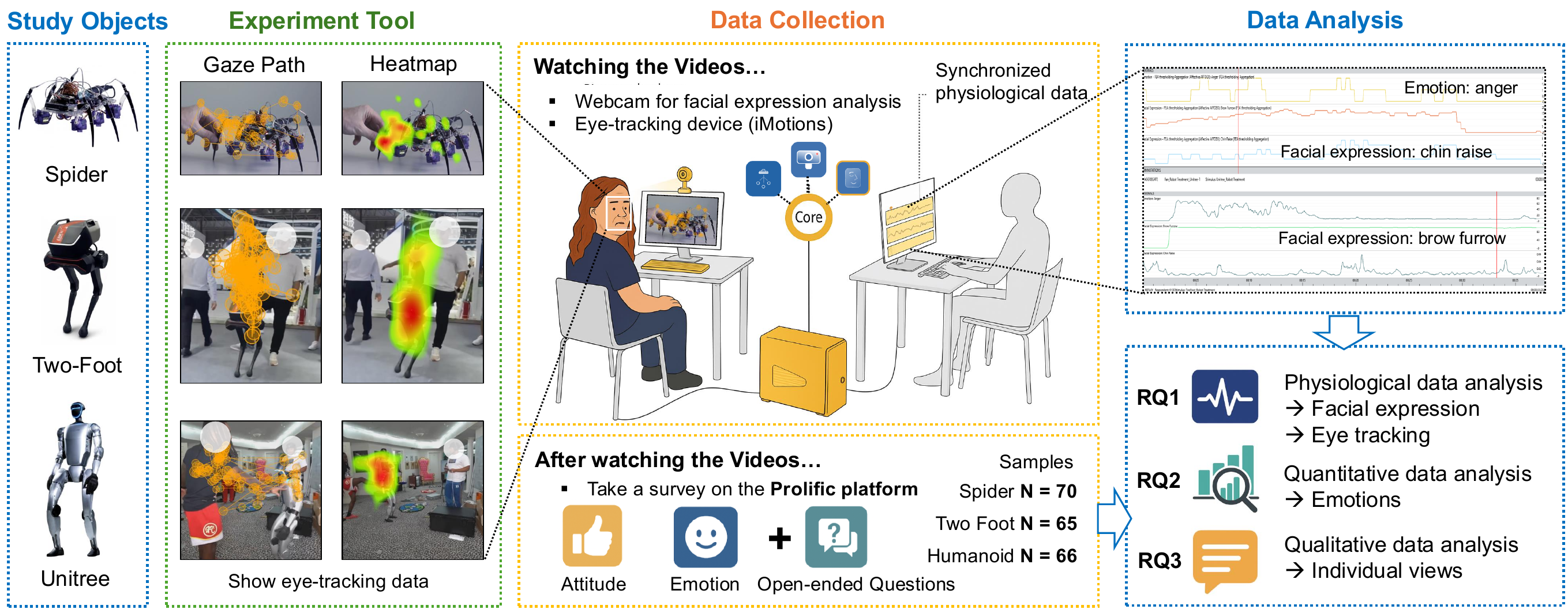}
    \caption{The illustrated framework for the implementation of our research.}
    \label{fig:framework}
\end{figure*}

\subsection{Research Stimuli}
The experimental stimuli were collected from publicly available videos showing users engaging in physical or verbal abuse toward three different robots that varied on the spectrum of anthropomorphism: a spider (Spider), a two-legged robot (Two-Foot), and a fully humanoid robot (Unitree).

The three videos were edited to have similar length (around 35 seconds) featuring abusive treatment of robots by humans. The spider robot video showed a human systematically breaking its legs one by one, the Two-Foot video depicted kicking and pushing that caused the robot to fall repeatedly, and the Unitree humanoid robot was subjected to strikes, shoves, and verbal taunts. While the specific behaviors necessarily varied according to each robot's morphology (breaking spider legs, kicking the bipedal robot, shoving the humanoid), each video depicted sustained, deliberate abuse resulting in functional impairment of the robot.

\subsection{Research Design and Participants}
This study employed a mix-methods experimental research design to investigate participants' reactions to and views of the abuse of robots across the spectrum of anthropomorphism (see Figure~\ref{fig:framework}). Upon IRB approval, participants (\textit{N}= 201) with an average age of 43.60 (\textit{SD} = 11.86, Males = 107, Females = 93) were recruited through Prolific, a high-quality research participant recruitment platform widely used by social scientists for understanding diverse human responses ~\cite{palan2018prolific, douglas2023data}. 
Participants accessed the study via a Qualtrics link on Prolific. After providing consent, they were randomly assigned to view one of three robot abuse videos while iMotions software ~\cite{boiangiu2022fast} tracked their facial expressions and eye movements through their webcams. Following the video, participants completed perceptions of the robot's uncanniness, subjective emotional response and open-ended questions regarding their thoughts of robot abuse (see Figure~\ref{fig:framework}).

\subsection{Measures and Data Analysis}

\subsubsection{Anthropomorphism}
Anthropomorphism was manipulated across three levels through video stimuli featuring robots with varying human-like characteristics: Spider (hexapod robot with mechanical limbs and no facial features), Two-Foot (bipedal robot with basic humanoid form but limited facial expressiveness), and fully humanoid (human-like appearance with expressive facial features and natural movements). This manipulation served as the independent variable for examining differential responses to robot abuse. A one-way ANOVA analysis was conducted to evaluate the manipulation and revealed that the three robots significantly differ in terms of their perceived humanness (Cronbach's alpha = .78, \textit{M} = 2.42, \textit{SD} = 1.14)---a key dimension of the uncanny valley ~\cite{ho2010revisiting,ho2017measuring}, \textit{F} (2,198) = 8.15, \textit{p} < .001. While there was an upward trend in perceived humanness from Spider (\textit{M} = 2.01, \textit{SE} = .11) to Unitree (\textit{M} = 2.76, \textit{SE} = .15), with Spider rated significantly lower than both other robots. Two-Foot's humanness (\textit{M} = 2.52, \textit{SE} = .14) fell in the middle—significantly higher than Spider, but similar to Unitree (See Figure~\ref{fig:perception_comparison}).

\subsubsection{Physiological Reactions}
Physiological reactions were captured via webcam using iMotions' Affectiva Affdex coding system ~\cite{kulke2020comparison}, which provided continuous measurement of discrete emotions through automated facial expression analysis. We specifically focused on anger expressions as a key indicator of moral outrage toward robot abuse (\textit{M} = .16, \textit{SD} = .891).

\subsubsection{Self-report Reactions}
\textbf{Robot Uncanniness}. Robot uncanniness was measured using the multi-facet scale by Ho and MacDorman's ~\cite{ho2017measuring, ho2010revisiting} with two relevant dimensions: erriness (Cronbach's alpha = .87, \textit{M} = 3.71, \textit{SD} = 1.40) and spine-tingling (Cronbach's alpha = .90, \textit{M} = 3.60, \textit{SD} = 1.28). In addition, participants' subjective \textbf{emotional reactions} were also measured using a multi-dimensional scale ~\cite{dillard2001persuasion} with a special focus on anger (\textit{M} = 3.77, \textit{SD} = 1.91) and guilt (\textit{M} = 3.09, \textit{SD} = 1.75) as key moral-related emotional reactions. Anger signals perceived injustice toward the robot (other-oriented moral concern), while guilt reflects discomfort with one's own passive witnessing of abuse (self-oriented moral concern). Unlike anger, which displays consistent facial markers ~\cite{stockli2018facial}, guilt lacks uniform facial expressions as it manifests primarily as an internal state, necessitating self-report measurement rather than automated facial coding.

\subsubsection{Open-Ended Responses to Robot Abuse}
Participants provided written responses describing their reactions to the robot abuse and their views on robot rights and ethical implications of robot protection. 

\subsubsection{Data Analysis}
We employed Multivariate Analysis of Variance (MANOVA) to examine physiological and self-report responses to robot abuse across varying levels of anthropomorphism (RQ1 and RQ2). To complement our quantitative findings, we conducted inductive analysis of qualitative data to explore participants' perspectives on the abuse of robots with varying degrees of anthropomorphism.

\section{RESULTS}

\subsection{Self-Report and Physiological Responses to Robot Abuse (RQ1 \& RQ2)}

A Multivariate Analysis of Variance (MANOVA) was conducted to 1) explore if perceived eeriness and spine-tingling---two dimensions of robot uncanniness ~\cite{ho2010revisiting} of the three robots---can vary as a function of their anthropomorphism per the prediction of the uncanny valley hypothesis, and 2) subsequently examine how individual responses to robot abuse vary by anthropomorphic levels of the robots.

\begin{figure}[t]
    \centering
    \includegraphics[width=0.6\linewidth]{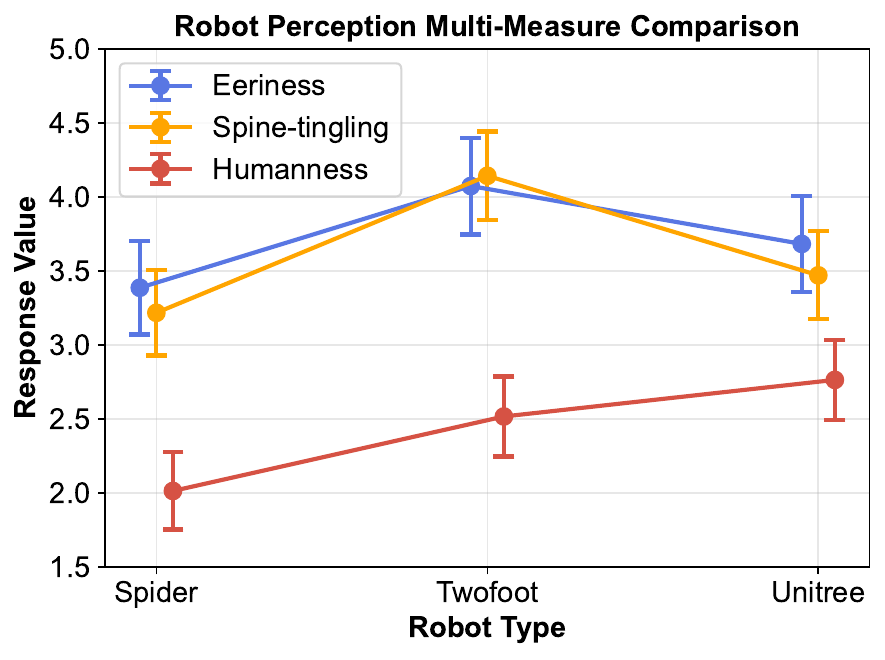}
    \caption{Multi-measure comparison of perceptions on the three robot types.}
    \label{fig:perception_comparison}
\end{figure}

This analysis uncovered a significant main effect of anthropomorphism, Wilks’ Lambda = .82, \textit{F} (10,388) = 4.14, \textit{p} < .001, partial eta squared = .10. The univariate analysis also indicated a significant main effect for self-report eeriness (\textit{F} (2,198) = 4.48, \textit{p} < .05) and spine-tingling (\textit{F} (2,198) = 10.17, \textit{p} < .001) (See Figure~\ref{fig:perception_comparison}). As shown in Table~\ref{tab:perception_comparison}, the perceived robot eeriness and spine-tingling indicate a clear uncanny valley, where near-humanness (Two-Foot) invoked significantly higher perceptions of uncanniness than both the least (Spider) and more (Unitree) anthropomorphic robots.

\begin{table}[htbp]
\centering
\caption{Pairwise comparisons on perceived eeriness, spine-tingling, and humanness}
\label{tab:perception_comparison}
\begin{threeparttable}
\begin{tabular}{@{}lcccc@{}}
\toprule
\textbf{Comparison} & \textbf{Mean Diff.} & \textbf{Std. Err.} & \textbf{Lower} & \textbf{Upper} \\
\midrule
\multicolumn{5}{@{}l@{}}{\textit{Eeriness}} \\
Spider vs Twofoot & $-0.687^{**}$ & 0.230 & $-1.141$ & $-0.233$ \\
Twofoot vs Unitree & $0.391$ & 0.233 & $-0.069$ & $0.852$ \\
Unitree vs Spider & $0.296$ & 0.229 & $-0.156$ & $0.748$ \\
\midrule
\multicolumn{5}{@{}l@{}}{\textit{Spine-tingling}} \\
Spider vs Twofoot & $-0.924^{***}$ & 0.211 & $-1.340$ & $-0.509$ \\
Twofoot vs Unitree & $0.672^{**}$ & 0.214 & $0.250$ & $1.093$ \\
Unitree vs Spider & $0.253$ & 0.210 & $-0.161$ & $0.666$ \\
\midrule
\multicolumn{5}{@{}l@{}}{\textit{Humanness}} \\
Spider vs Twofoot & $-0.503^{**}$ & 0.190 & $-0.878$ & $-0.127$ \\
Twofoot vs Unitree & $-0.247$ & 0.193 & $-0.628$ & $0.134$ \\
Unitree vs Spider & $0.749^{***}$ & 0.190 & $0.375$ & $1.123$ \\
\bottomrule

\end{tabular}
\begin{tablenotes}
\item Note: 95\% confidence intervals shown using Holm’s sequential bonferroni post hoc comparisons. $^{*}p<.05$, $^{**}p<.01$, $^{***}p<.001$.
\end{tablenotes}
\end{threeparttable}
\end{table}

\begin{figure}[t]
    \centering
    \includegraphics[width=0.6\linewidth]{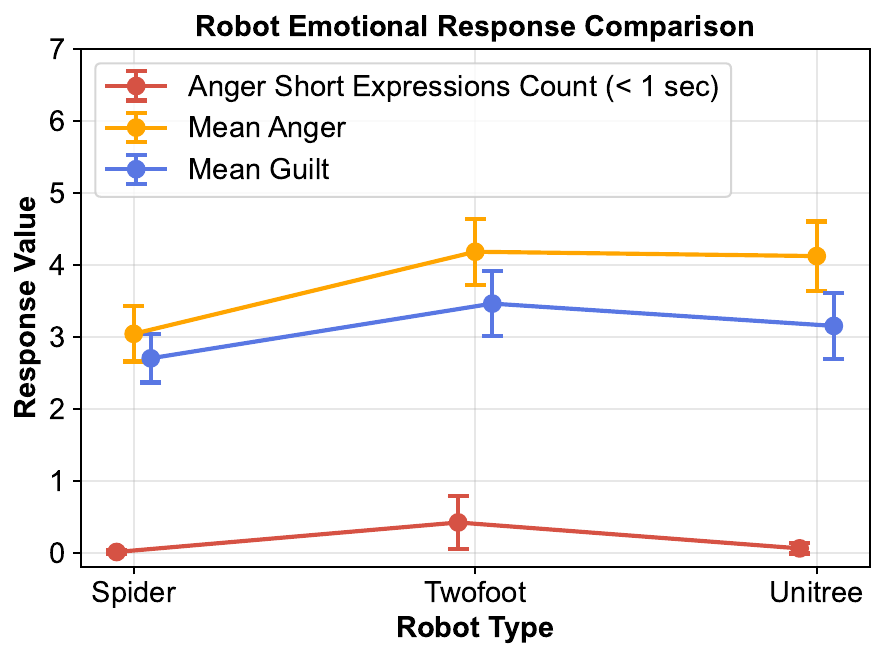}
    \caption{Comparison of emotional responses to the three robot types.}
    \label{fig:emotion_comparison}
\end{figure}

 The univariate analysis also indicated a significant main effect for self-report anger, \textit{F} (2,198) = 8.30, \textit{p} < .001, partial eta squared = .08. Two-Foot  (\textit{M} = 4.19, \textit{SE} = .23) and Unitree (\textit{M} = 4.12, \textit{SE} = .23) triggered significantly higher self-report anger than Spider (\textit{M} = 3.04, \textit{SE} = .22) with no significant difference between them. In addition, a significant main effect for self-report guilt also emerged, \textit{F} (2,198) = 3.33, \textit{p} < .05, partial eta squared = .03. Similarly, Two-Foot (\textit{M} = 3.46, \textit{SE} = .21) and Unitree (\textit{M} = 3.15, \textit{SE} = .21) provoked participants' feeling of guilt significantly higher than Spider did (\textit{M} = 2.70, \textit{SE} = .21) (See Table~\ref{tab:emotion_comparison} and Figure~\ref{fig:emotion_comparison}).

The same multivariate analysis also indicated a significant  main effect of anthropomorphism on participants facially expressed anger. The univariate analysis indicated a significant main effect for physiological anger expression for anthropomorphism (\textit{F} (2, 198) = 4.14, \textit{p} < .05, partial eta squared = .04), with Two-Foot (\textit{M} = .42, \textit{SE} = .11) invoking significant more anger expressed in participants facial expressions than Spider (\textit{M} = .01, \textit{SE} = .11) and Unitree (\textit{M} = .06, \textit{SE} = .11) (See Table~\ref{tab:emotion_comparison} and Figure~\ref{fig:emotion_comparison}). 

\begin{table}[htbp]
\centering
\caption{Pairwise Comparisons on Physiological and Self-report Reactions}
\label{tab:emotion_comparison}
\begin{threeparttable}
\begin{tabular}{@{}lcccc@{}}
\toprule
\textbf{Comparison} & \textbf{Mean Diff.} & \textbf{Std. Err.} & \textbf{Lower} & \textbf{Upper} \\
\midrule
\multicolumn{5}{@{}l@{}}{\textit{Anger (irritated, angry, annoyed, aggravated)}} \\
Spider vs Twofoot & $-1.142^{***}$ & 0.317 & $-1.767$ & $-0.516$ \\
Twofoot vs Unitree & $0.063$ & 0.322 & $-0.571$ & $0.698$ \\
Unitree vs Spider & $1.078^{***}$ & 0.316 & $0.455$ & $1.702$ \\
\midrule
\multicolumn{5}{@{}l@{}}{\textit{Guilt (guilty, ashamed)}} \\
Spider vs Twofoot & $-0.762^{*}$ & 0.298 & $-1.349$ & $-0.175$ \\
Twofoot vs Unitree & $0.310$ & 0.302 & $-0.286$ & $0.906$ \\
Unitree vs Spider & $0.452^{+}$ & 0.297 & $-0.133$ & $1.036$ \\
\midrule
\multicolumn{5}{@{}l@{}}{\textit{Anger Short Expressions Count ($<$ 1 sec)}} \\
Spider vs Twofoot & $-0.401^{**}$ & 0.151 & $-0.699$ & $-0.103$ \\
Twofoot vs Unitree & $0.355^{*}$ & 0.153 & $0.052$ & $0.657$ \\
Unitree vs Spider & $0.046$ & 0.151 & $-0.251$ & $0.343$ \\
\bottomrule
\end{tabular}
\begin{tablenotes}
\item Note: 95\% confidence intervals shown using Holm’s sequential bonferroni post hoc comparisons. $^{+}p<.10$, $^{*}p<.05$, $^{**}p<.01$, $^{***}p<.001$. 
\end{tablenotes}
\end{threeparttable}
\end{table}

\subsection{Individual Views on Robot Abuse and Protection across Anthropomorphism Levels (RQ3)}

Our participants' responses reveal that a robot's degree of anthropomorphism largely shapes their perception of its mistreatment in three ways, including their immediate reactions (Sec~\ref{sec:empathy_overview}), moral reasoning (Sec~\ref{sec:moral_justifications}), and proposals for social implications (Sec~\ref{sec:governance_overview}). As a robot’s appearance shifts from mechanical to life-like, its perception moves from viewing its ``abuse'' as a technical or property issue to viewing it as a complex social and ethical problem. Table~\ref{tab:findings_summary} summarizes these findings.

\subsubsection{Empathy for Robots Increases in Direct Proportion to Their Anthropomorphism}
\label{sec:empathy_overview}
\mbox{}\

\noindent We found that as a social robot’s design becomes more anthropomorphic, participants tended to respond with strong empathy, project feelings onto the robot, and morally condemn the humans’ actions in the stimuli videos.

\textbf{The Spider Robot: Technical Assessment with Irrational Empathy.}
For the least anthropomorphic robot, a six-legged ``spider,'' participants often reacted with a technical assessment of the robot's capabilities. Many focused on such robots' mechanical properties, with one participant commenting they (people in the video) were \textit{``watching to see if the robot adapted to having shorter legs.''} Another viewed it as \textit{``excellent training on how a robot can still function despite damage.''} This lens was frequently used to justify a lack of emotion, leading participants to conclude that since the robot \textit{``can't feel any pain, so it's no big deal.''}

However, even with this detachment, many participants expressed a form of qualified or \textit{``irrational''} empathy, often explicitly labeling their own emotional response as surprising or illogical. This suggested a conflict between their rational understanding and their gut feeling, as one participant noted, \textit{``Even though the robot is not real and has no feelings, it made me feel a bit sad for some reason.''} Another described this internal conflict clearly:
\begin{quote}
\textit{I did irrationally feel bad for it. I was affected in the way one might torture a spider or crab for one's amusement.}
\end{quote}

This self-conscious empathy was unique to the spider robot, suggesting its non-humanoid form created a cognitive dissonance where feeling bad for the ``machine'' required justification.

\textbf{The Two-Foot Robot: Interpreting the Action as Social Bullying.}
The humanoid robot's form prompted participants to interpret the action through the lens of social bullying, projecting human social dynamics onto the action. Participants overwhelmingly cast the robot in the role of a defenseless human victim, with one stating, \textit{``At first it seemed a bit sad. I imagined a small child being bullied.''} This differed from the irrational empathy shown toward the spider robot, as the emotional distress here was a natural and direct response to a perceived social transgression. Participants commented that it was \textit{``painful to watch''} and that it was difficult \textit{``not to think of the robot as a living creature.''}

Flowing from this reaction, participants engaged in a critique of human cruelty. The focus here shifted from the robot's experience to a condemnation of the human actors' character. Their behavior was seen as a reflection of negative aspects of human nature, prompting participants' comments like, \textit{``It made me think of how cruel humans can act,''} and expressions of \textit{``Anger at the way people react to things that cannot fight back.''}

\textbf{The Unitree Robot: A Mix of Animal-like Empathy and Moral Disgust.}
The animal-like, four-legged Unitree robot elicited a mix of animal-like empathy and moral disgust. The projection of vulnerability was distinctly zoomorphic, as participants frequently compared the robot to a defenseless animal or child being tormented. This reaction elicited feelings of protectiveness and guilt as a participant explained:
\begin{quote}
\textit{I felt guilty for how it was being treated, like it was defenseless and somewhat like a child or animal being misled into violence.}
\end{quote}

This empathy was often accompanied by moral disgust and shame directed at the humans, as the judgment of their character was often visceral. Participants stated the scene was \textit{``sad and disgusting''} or that they \textit{``felt awful for the robot and ashamed to be a part of the human race.''}

\begin{table}[t]
\centering
\small 
\caption{A summary of Primary Themes that Emerged across the Three Robot Morphologies}
\label{tab:findings_summary}
\begin{tabularx}{\columnwidth}{>{\raggedright\arraybackslash}m{2.3 cm} >{\raggedright\arraybackslash}X >{\raggedright\arraybackslash}X >{\raggedright\arraybackslash}X}
\toprule
\textbf{Robot Morphology} & \textbf{Immediate Reactions} & \textbf{Moral Reasoning} & \textbf{Social Implications} \\
 & \textit{(Sec~\ref{sec:empathy_overview})} & \textit{(Sec~\ref{sec:moral_justifications})} & \textit{(Sec~\ref{sec:governance_overview})} \\
\midrule
\textbf{\raisebox{-.5\height}{\includegraphics[height=1cm]{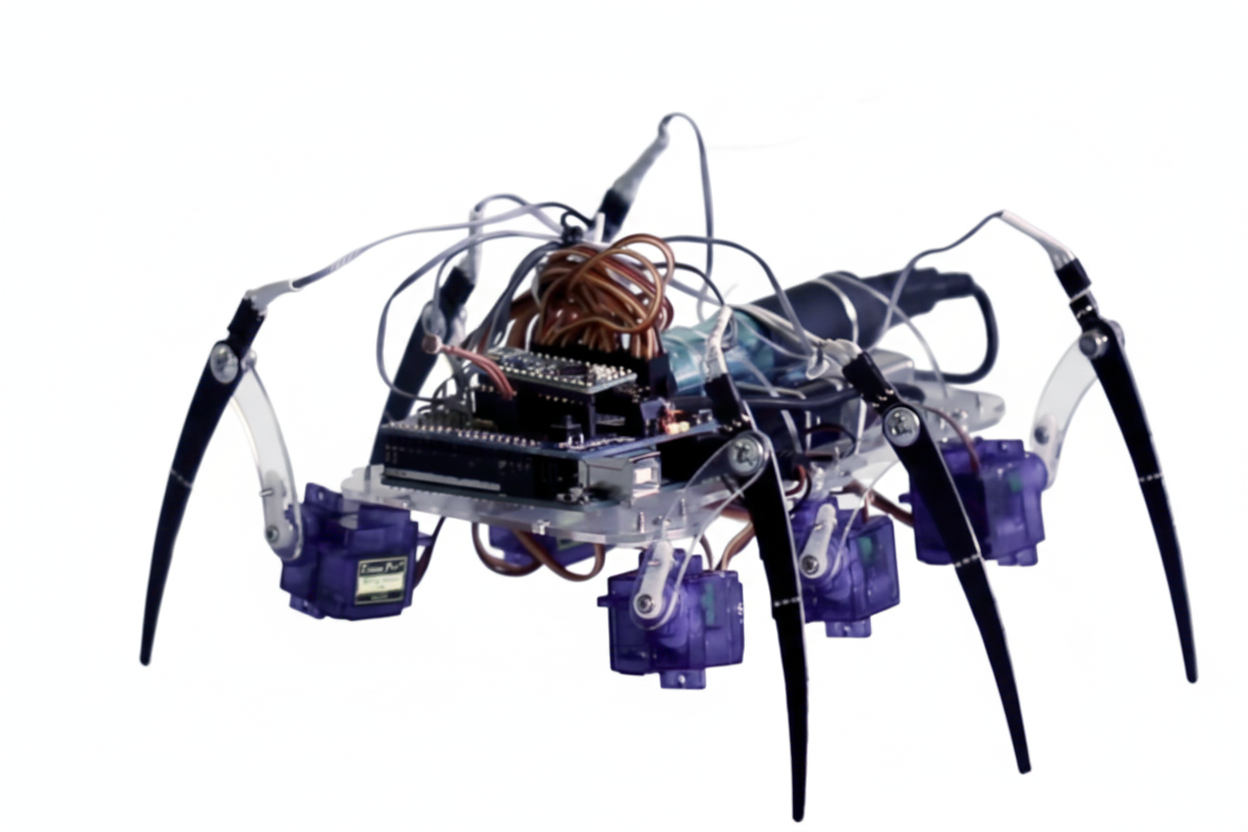}}\hspace{0.5em}Spider} & Technical Assessment with Irrational Empathy & A Question of Property and Sentience & A Debate Centered on Property Law \\ \addlinespace
\textbf{\raisebox{-.5\height}{\includegraphics[height=1.1cm]{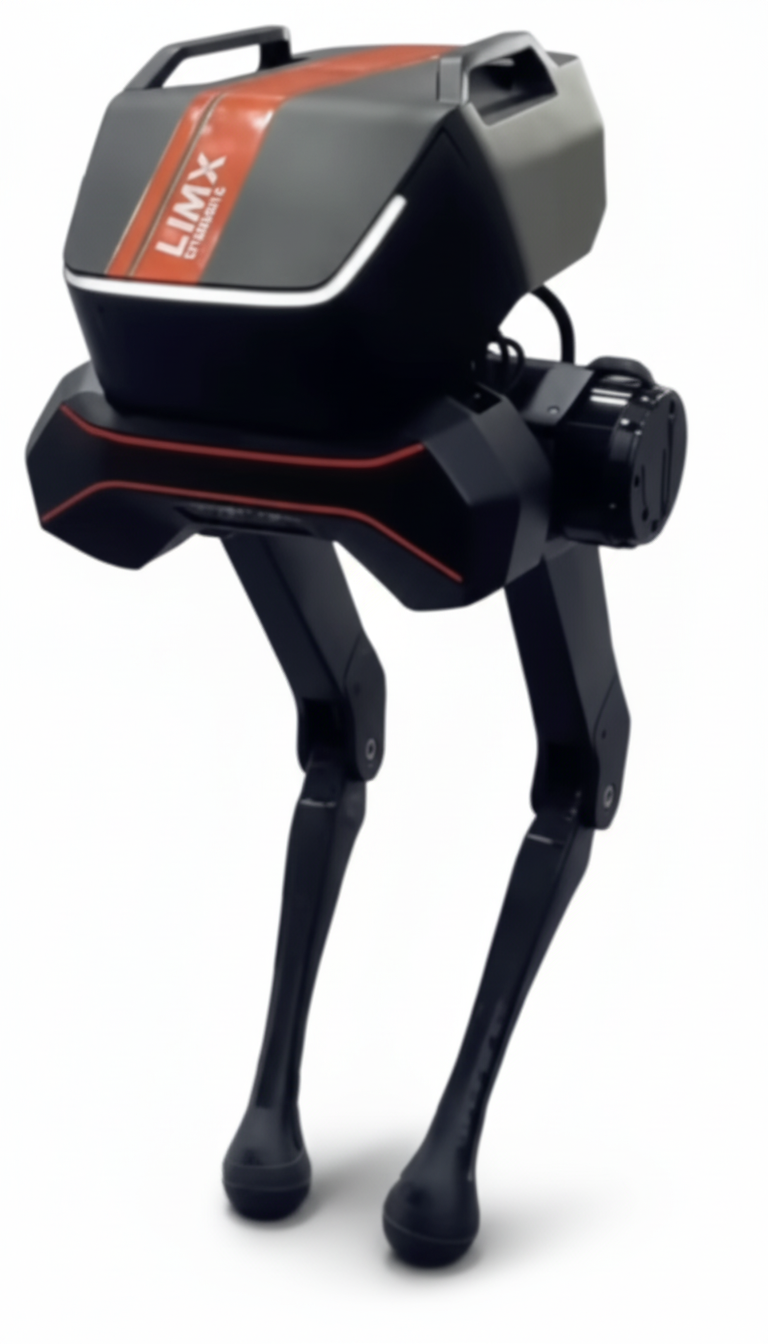}}\hspace{0.5em}Two-Foot} & Interpreting the Action as Social Bullying & A Reflection of Human Character & A Shift Towards Shared Responsibility \\ \addlinespace
\textbf{\raisebox{-.5\height}{\includegraphics[height=1.1cm]{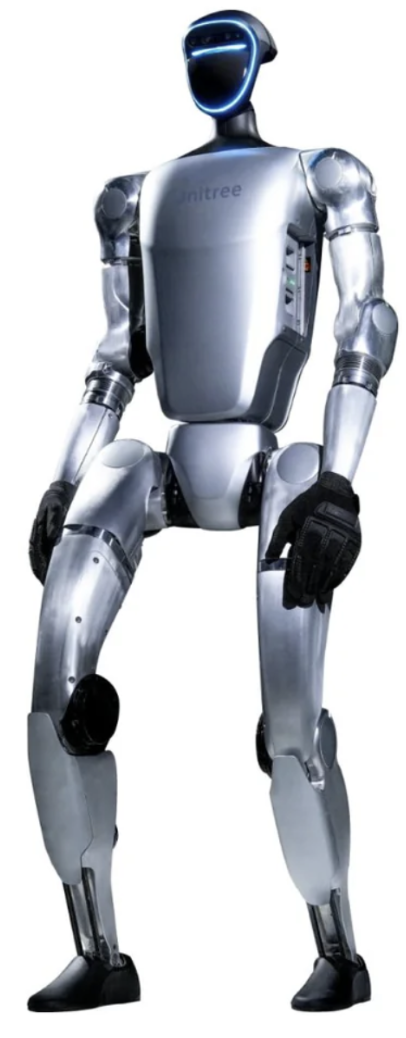}}\hspace{0.5em}Unitree} & A Mix of Animal-like Empathy and Moral Disgust & A Proxy for a Living Being & A Debate Between Quasi-Animal Rights and Object Status \\
\bottomrule
\end{tabularx}
\end{table}

\subsubsection{Moral Justifications Shift from Property Rights to Human Character as Anthropomorphism Increases}
\label{sec:moral_justifications}
\mbox{}\

\noindent Beyond immediate reactions, participants’ moral reasoning also varied with the robot's morphologies. For the less life-like robot, the debate was about property and sentience, while as anthropomorphism increased, participants' moral reasoning shifted to concerns about human character and the potential for real-world harm.

\textbf{The Spider Robot: A Question of Property and Sentience.}
\label{sec:spider_moral}
For the six-legged spider robot, the moral debate centered on its status as an inanimate object and a question of property and sentience. The most common justification for the treatment being acceptable was the robot's lack of sentience, as participants argued that since it was just a \textit{``machine with no feelings,''} human concepts of morality did not apply. This perspective framed the robot as no different from any other mechanical object, with one participant stating, \textit{``There is no violation of morals for destroying a mechanical item (think of the people who smash their cellphone out of anger).''}

However, even among those who viewed the robot as mere property, some participants still found the action unacceptable, not because the robot had rights, but because its destruction was \textit{``wasteful and disrespectful to destroy property.''} This view bases the moral wrongness not in the robot's experience, but in the violation of principles regarding robot property and utility.




\textbf{The Two-Foot Robot: A Reflection of Human Character.}
The humanoid robot's form complicated participants' moral reasoning, shifting the focus from object status to a reflection of human character. While many still maintained that the treatment was acceptable because the robot \textit{``does not have consciousness,''} a significant portion felt it was morally wrong precisely because of its life-like qualities. Its human-like movement and struggles made the abuse feel wrong, even if participants intellectually understood it was a machine, with one explaining this conflict: \textit{``since it seems so life-like, I do feel like it is wrong to treat it abusively.''}

This led participants to reason that the action served as a moral litmus test for humans. The robot's treatment was seen not as an isolated act, but as a direct indicator of a person's underlying character and their potential to harm others. One participant explained this perspective:
\begin{quote}
\textit{I do however think that a person who wants to treat a robot poorly likely also would want to treat people poorly.}
\end{quote}

This perspective shifts the moral focus entirely away from the robot's experience and onto the human actor. The robot's lack of sentience becomes irrelevant because, as another participant argued, the act of \textit{``bullying''} something that \textit{``cannot fight back is a moral failing.''} This reasoning suggests that such behavior reveals a disposition for cruelty that could easily \textit{``spill over elsewhere''} to sentient beings, making the act itself a sign of poor character.

\textbf{The Unitree Robot: A Proxy for a Living Being.}
For the animal-like Unitree robot, the moral lines blurred the most, as many participants judged the action by treating the robot as a proxy for a living being. Many abandoned the machine/object distinction entirely, arguing that the treatment was equivalent to animal abuse. The robot's zoomorphic appearance and movements created a strong empathetic connection, leading many participants to conclude the act was \textit{``totally wrong and akin to torturing an animal.''} The moral judgment was often instinctual, with participants stating it \textit{``felt wrong even if it isn't human.''}

While some participants firmly held that morality was irrelevant because it was \textit{``just a machine with no feelings,''} a larger portion engaged with the action as a reflection of human morality. As with the Two-Foot robot, participants invoked the slippery slope argument, but often with more urgency. One participant commented:
\begin{quote}
\textit{No I don't think it is because if you could treat a mechanical object that resembles a human like this then you'll treat living things like this. It's like a slippery slope and those don't end well.}
\end{quote}

This reasoning connects the treatment of an animal-like robot directly to future actions toward living beings, framing the abuse not just as an indicator of character but as a potential training ground for real-world harm.

Finally, some participants framed the issue as a test of positive character. From this perspective, choosing not to harm the robot, even if it feels nothing, is a mark of a good person. The act was seen as wrong because it is \textit{``disrespectful to your own character/soul to emulate a bully that way''} and because it is wrong \textit{``to get enjoyment out of harming something that can't harm you back.''}

\subsubsection{Proposals for Governance Are Directly Shaped by a Robot's Anthropomorphism}
\label{sec:governance_overview}
\mbox{}\

\noindent Beyond personal moral judgments, participants’ proposals for societal rules and legal protections also varied with the robot's form. For the less life-like robot, the debate centered on property law, while for more anthropomorphic robots, it expanded to include shared responsibility and quasi-animal rights.

\textbf{The Spider Robot: A Debate Centered on Property Law.}
For the six-legged spider robot, the conversation about social rules was grounded in established legal concepts, leading to a debate centered on property law. The primary perspective here was that no new laws are needed because the robot is fundamentally an object. Participants argued that special protections were unnecessary for a machine that is \textit{``not real and they do not have any emotions or thoughts.''} From this, existing laws are sufficient because the robot is already covered as personal property. As one participant stated: \textit{``Robots already have protection. It's called `property law'.''} This case reduces the complex ethical questions to a simple, pre-existing legal category, thereby dismissing the need for any new moral or legal considerations specific to robots. This was a common view, with another participant stating that the \textit{``same laws that apply to other property should apply to robots.''}

However, a strong counter-argument emerged that focused not on the robot's welfare, but on the social implications of the human's actions. This group argued for regulations as a way to police worrying human behavior, expressing concern that cruelty towards an inanimate object could be a precursor to violence against sentient beings. One participant noted that:
\begin{quote}
\textit{…mistreating a robot opens the door for a human to depersonalize a living being and we should not train people to depersonalize.}
\end{quote}

Here, the reasoning for legal protection is reframed entirely; it is not about granting rights to the robot, but about preventing the normalization of humans' cruel behavior for the protection of society at large.

\textbf{The Two-Foot Robot: A Shift Towards Shared Responsibility.}
The humanoid robot's appearance prompted a forward-looking debate about governance, causing a shift towards shared responsibility. Many participants took a conditional stance, arguing that while special laws are not necessary now, they may be required in the future if technology continues to advance. 

Unlike the spider robots, participants in this condition moved beyond a simple yes/no debate to construct a framework of shared responsibility. They envisioned a multi-stakeholder approach where no single entity was solely responsible, involving individuals, companies, and the government. As a participant explained, \textit{``I would say that it requires a combination of the aforementioned stakeholders. This cannot fall on just one segment of society.''} This view suggests that as robots become more human-like, the governance of their treatment becomes a complex societal project, with another participant concluding that \textit{``the people producing the robot and the government''} share an obligation because the government \textit{``sets the tone for how we should act toward them.''}

\textbf{The Unitree Robot: A Debate Between Quasi-Animal Rights and Object Status.}
The animal-like Unitree robot elicited the most polarized arguments for and against legal protection, creating a debate between quasi-animal rights and object status. The discussion was no longer primarily about property but about the robot's fundamental moral standing. One side of participants argued that the robot deserves protection, often equating its treatment with that of a living being. Participants commented that robots should be treated with \textit{``dignity and respect, just like a human being''} and should have protections \textit{``against harm from others and the right to exist.''} This view moves far beyond property law and toward granting the robot a form of moral patienthood. A participant suggested:
\begin{quote}
\textit{Yes, absolutely. The robot should be able to defend itself from such abuse, or criminal action should be taken.}
\end{quote}

This case assigns the robot a right to self-defense, a concept typically reserved for living beings, demonstrating how deeply the robot's animal-like form blurred the line between machine and moral entity.

In direct opposition was the theme of rejection of robot rights, which dismissed the idea that a machine could or should have rights. As a participant argued:
\begin{quote}
\textit{No. You wouldn't give protection to a toaster, would you? Just because the toaster now resembles a human being doesn't mean it's human. It's still a toaster.}
\end{quote}

This argument rejects anthropomorphism as a basis for rights, anchoring the robot's status in its mechanical origin, regardless of its appearance or behavior. Proponents of this view often found the discussion itself absurd, stating that protections are \textit{``ridiculous to give a machine rights''} or worrying that society has \textit{``lost concept of living beings.''} 

\section{DISCUSSION}
\subsection{No Valley in Protection despite Uncanny Perception}
Consistent with the original uncanny valley hypothesis, we observed a clear valley effect in perceived eeriness and spine-tingling sensations, with the intermediate anthropomorphic robot (Two-Foot) triggering significantly higher uncanniness than both the mechanical Spider and the more sophisticated Unitree (Table 1). However, this perceptual uncanniness failed to suppress empathic or protective responses. Both Two-Foot and Unitree elicited similarly elevated levels of self-reported anger and guilt compared to Spider, with no significant difference between them despite their divergent uncanniness ratings (Table~\ref{tab:emotion_comparison}).

Interestingly, Two-Foot—the robot rated most uncanny—actually generated the strongest physiological anger expressions, significantly exceeding both Spider and Unitree, creating what might be characterized as an inverted uncanny valley in protective responses. Two-Foot's liminal level of humanness between machine and human could have created a special moral standing for robots that are human-like but not quite human enough—triggering participants to react with heightened physiological activation, as they cannot easily rely on existing social scripts that typically guide human-computer interaction (as CASA suggests ~\cite{nass1993voices, nass2000machines}) to evaluate its abuse (neither fitting animal abuse nor human violence frameworks), forcing more effortful and elaborative processing that manifests physiologically even without conscious awareness.

This dissociation between perceptual uncanniness and moral considerations becomes even clearer when examining participants' reasoning about robot protection and governance. Despite Two-Foot's peak uncanniness, it still prompted sophisticated discussions about shared responsibility and the need for multi-stakeholder governance (Section ~\ref{sec:governance_overview}). Moreover, participants argued that treatment of Two-Foot served as a moral test (Section~\ref{sec:moral_justifications})---framing the issue not as one of robot rights but of human character and potential for harm toward sentient beings. Clearly, the uncanny valley became irrelevant when it comes to conscious moral deliberations; participants focused on what the abuse revealed about human nature and its societal implications rather than their aesthetic discomfort with the robot's appearance.

\subsection{The Power of Anthropomorphism in Robot Protection}
Our findings highlight the role served by anthropomorphism as the primary driver of protective responses - a notable departure from the uncanny valley effect. The qualitative data reveal a clear progression in moral engagement that correlates directly with anthropomorphic features. While Spider's mechanical form elicited what participants explicitly labeled as ``irrational'' empathy—a self-conscious response they felt compelled to justify despite the robot's non-humanlike appearance—the more anthropomorphic Two-Foot and Unitree triggered immediate, unquestioned moral responses, with participants invoking social concepts like ``social bullying,'' and ``cruelty'' (Section~\ref{sec:empathy_overview}).

This anthropomorphism-driven shift aligns well with the quantitative findings regarding perceived humanness—Spider rated significantly lower than both Two-Foot and Unitree, which were perceived as similarly human-like. This pattern manifests strikingly in the qualitative data: Spider elicited explicitly ``irrational'' empathy that participants felt compelled to justify (Section~\ref{sec:empathy_overview}) by focusing on technical assessment and property concerns (Section~\ref{sec:moral_justifications}). In contrast, both Two-Foot and Unitree triggered participants to frame them as victims through human or animal analogies (Section~\ref{sec:empathy_overview}).

This binary divide extends to governance proposals, where Spider prompted property law debates, while both Two-Foot and Unitree elevated discussions to societal implications—Two-Foot generating calls for shared responsibility among ``individuals, companies, and governments'' and Unitree sparking debates about whether robots deserve ``dignity and respect, just like a human being'' (Section~\ref{sec:governance_overview}). The convergence of responses to Two-Foot and Unitree, despite their different forms but similar perceived humanness, demonstrates that anthropomorphism operates as a governing factor driving up human protective responses to robot abuse.

\subsection{Theoretical Implications}
This study contributes to the existing CASA and uncanny valley literature by empirically revealing important boundary conditions for anthropomorphism, demonstrating that increasing humanness, albeit at risk of triggering perceptual eeriness as predicted by the uncanny valley effect ~\cite{mori2012uncanny}, does not transfer to suppress moral concern or protective responses toward robot abuse. Instead, we find an inverted uncanny valley in protective responses—the uncanny Two-Foot generated the strongest physiological anger, suggesting that categorical ambiguity in near-humanness may act as a moral amplifier rather than inhibitor. This extends CASA theory ~\cite{nass1993voices, nass2000machines} by revealing that while people typically mindlessly or mindfully apply existing social scripts to social agents, ambiguous humanness may disrupt both script activation and application processes, forcing more effortful responses when no clear framework (neither human violence nor animal cruelty nor property damage) readily applies. Furthermore, our findings demonstrate that anthropomorphism fundamentally drives moral reasoning in robot abuse. These insights challenge both uncanny valley and CASA theories to account for contexts where moral stakes outweigh perceptual comfort, advancing our understanding of human-robot interaction specifically in moral contexts.

\subsection{Design and Policy Implications}
The central role of anthropomorphism in promoting protective responses, as revealed in this study, suggests that designers should prioritize human-like features when designing robots that are primarily deployed in contexts with heightened risk of abuse—such as public spaces, service industries, or educational settings ~\cite{shiomi2025robot, shum2024kicking, tan2018inducing, sun2026frontline}. Even when a robot has to be designed with only partial human features due to its specific application, it may still elicit strong protective responses than clearly mechanical designs, as suggested by Two-Foot's heightened physiological responses.

Policy implications from our findings point to the need for differentiated ethical frameworks that recognize anthropomorphism's governing role in protective responses to robot abuse. The starkly different reactions found in our study suggest a targeted regulatory approach: for minimally anthropomorphic robots like Spider, which triggered property-focused concerns, regulations should focus on preventing wasteful destruction and property damage. However, for robots with clear anthropomorphic features---as both Two-Foot and Unitree did despite their different forms---participants envisioned multi-stakeholder governance involving manufacturers, governments, and users, indicating the need for comprehensive protective frameworks beyond simple property law.

\subsection{Limitations \& Future Work}

Our study presents several opportunities for future work. While our study includes 201 participants recruited through Prolific, the sample may not fully represent the diversity of demographic or experiential backgrounds that shape people’s moral intuitions toward robots. Online recruitment platforms attract individuals with varying familiarity with technology, but they often skew toward young and educated populations. As responses to anthropomorphism and moral concern may be culturally contingent, future research could test whether the observed patterns generalize across different demographic groups and levels of technological exposure. In addition, although our mixed-methods design combined surveys, physiological measures, and open-ended reflections, these data are still collected in an online setting rather than through lived encounters with robots in physical environments.

Another limitation is the scope of robots examined. Our study focuses on three types that may not represent the full diversity of robotic forms people encounter in everyday life. Social, service, companion, and industrial robots all incorporate distinct design features, behavioral cues, and contexts of use that could differently shape protective responses. Exploring a broader range of robot morphologies, functions, and interaction scenarios would clarify whether our findings extend beyond the specific cases studied. For instance, animal-inspired robots, voice-only agents, or hybrid embodied systems may elicit different blends of empathy, discomfort, or moral concern. Future work could therefore expand the design space to better map how anthropomorphism interacts with context to influence protective impulses.

\section{CONCLUSIONS}
This study utilizes mixed-methods to examine protective responses to robot abuse. While the uncanny valley emerges in perceptions of robots, it fails to influence protective responses. Anthropomorphism fundamentally powers moral reasoning regarding robot abuse and governs protective responses independent of the uncanny effect. As social robots become prevalent, these findings underscore the need for anthropomorphism-calibrated design and policy approaches that recognize how human-like features, not aesthetic (dis)comfort, determine moral considerations particularly for robots at high risks of abuse.


\bibliographystyle{ACM-Reference-Format}
\bibliography{main}

\appendix



\section{Survey Design}

The following table details the open-ended questions presented to participants in the qualitative survey. The survey was structured to first gather responses on responsibility attribution before probing opinions on the continuity of service.

\small
\begin{longtable}{@{}p{0.95\textwidth}@{}}
\toprule
\textbf{Robot Abuse — Survey Questions} \\
\midrule
\endfirsthead
\toprule
\textbf{Robot Abuse — Survey Questions (continued)} \\
\midrule
\endhead

\textbf{Start of Block: Prolific ID} \\
\textbf{Prolific ID} What is your Prolific ID?  Please note that this response should auto-fill with correct ID \\
\rule{\linewidth}{0.4pt}\\[0.8em]

\textbf{Start of Block: Transition} \\
\textbf{Transition} Now you are going to watch a video of a robot being treated in various ways. Please watch the video in its entirety carefully. Your eye movement and facial expressions will be recorded while you watch the video. \\
\addlinespace[1em]

\textbf{Start of Block: Emotion} \\[0.25em]

\textbf{Emotions} When watching the video, I feel \\
\emph{Strongly disagree (1), Disagree (2), Somewhat disagree (3), Neither agree nor disagree (4), Somewhat agree (5), Agree (6), Strongly agree (7)} \\[0.5em]

Surprise (surprised, startled, astonished) (1) [Emotion 1–7] \\
Anger (irritated, angry, annoyed, aggravated) (2) [Emotion 1–7] \\
Fear (fearful, afraid, scared) (3) [Emotion 1–7] \\
Sadness (sad, dreary, dismal) (4) [Emotion 1–7] \\
Guilt (guilty, ashamed) (5) [Emotion 1–7] \\
Happiness (happy, elated, cheerful, joyful) (6) [Emotion 1–7] \\
Contentment (contented, peaceful, mellow, tranquil) (7) [Emotion 1–7] \\
\addlinespace[1em]

\textbf{Start of Block: Uncanny Valley} \\[0.25em]

\textbf\;Please rate your impression of the robot you saw in the video: \\
\emph{(1)\; (2)\; (3)\; (4)\; (5)\; (6)\; (7)} \\[0.4em]

Dull — Freaky \; [Uncanny Valley 1–7] \\
Predictable — Eerie \; [Uncanny Valley 1–7] \\
Plain — Weird \; [Uncanny Valley 1–7] \\
Ordinary — Supernatural \; [Uncanny Valley 1–7] \\
Boring — Shocking \; [Uncanny Valley 1–7] \\
Uninspiring — Spine-tingling \; [Uncanny Valley 1–7] \\
Predictable — Thrilling \; [Uncanny Valley 1–7] \\
Bland — Uncanny \; [Uncanny Valley 1–7] \\
Unemotional — Hair-raising \; [Uncanny Valley 1–7] \\
Inanimate — Living \; [Uncanny Valley 1–7] \\
Synthetic — Real \; [Uncanny Valley 1–7] \\
Mechanical Movement — Biological Movement \; [Uncanny Valley 1–7] \\
Human-Made — Humanlike \; [Uncanny Valley 1–7] \\
Without Definite Lifespan — Mortal \; [Uncanny Valley 1–7] \\
\addlinespace[1em]

\textbf{Start of Block: Open-ended Questions} \\[0.25em]

\textbf{Immediate Reactions} What went through your mind when you watched how the robot was treated? \\

\textbf{Uncanny Valley} On a spectrum from machine to human, where would you place this robot you just saw in the video? Please explain your reasoning \\

\textbf{Moral Reasoning1} In your view, is it morally acceptable to treat the robot as it was in the video? Please explain your reasoning. \\

\textbf{Moral Reasoning2} Now, consider the same actions directed toward an animal. How would your reaction differ, and why? \\

\textbf{Social Implications1} Do you think society should have rules or laws about how people treat robots, and should robots be given legal protection? If so, what kinds of rights or protections do you think would be important? Please explain your reasoning. \\

\textbf{Social Implications2} Who do you think should be responsible for ensuring that robots are treated ethically - individuals, companies that make robots, the government, or someone else? Why? \\

\bottomrule
\end{longtable}

\twocolumn
\clearpage

\end{document}